%% file: cvpr.tex
\documentclass[review]{cvpr}

\usepackage{times}
\usepackage{epsfig}
\usepackage{graphicx}
\usepackage{amsmath}
\usepackage{amssymb}
\usepackage{chngpage}
\usepackage{mathtools}
\usepackage{caption}
\usepackage[table,xcdraw]{xcolor}
\usepackage[section]{placeins}
\usepackage{multirow}
\usepackage{listings}

\usepackage[pagebackref=true,breaklinks=true,colorlinks,bookmarks=false]{hyperref}

\begin{document}

\title{UFO-ViT: High Performance Linear Vision Transformer without Softmax}

\author{
Jeong-geun Song\\
Kakao Enterprise\\
{\tt\small po.ai@kakaoenterprise.com}
}

\maketitle
\begin{abstract}
Vision transformers have become one of the most important models for computer vision
tasks. Although they outperform prior works, they require heavy computational resources on a scale that is quadratic to $N$. This is a major drawback of the traditional self- attention (SA) algorithm. Here, we propose the Unit Force Operated Vision Transformer (UFO-ViT), a novel SA mechanism that has linear complexity. The main approach of this work is to eliminate nonlinearity from the original SA. We factorize the matrix multiplication of the SA mechanism without complicated linear approximation. By modifying only a few lines of code from the original SA, the proposed models outperform most transformer-based models on image classification and dense prediction tasks on most capacity regimes.
\end{abstract}

\input{section/introduction}
\input{section/related_works}
\input{section/methods}
\input{section/experiments}
\input{section/conclusion}
\bibliographystyle{ieee_fullname}
\bibliography{cvpr.bbl}
\clearpage
\end{document}

%% file: section/introduction.tex
\section{Introduction}
As early successes in natural language processing (NLP), several studies based on
transformers \cite{dosovitskiy2020image, touvron2020training, wu2021cvt, srinivas2021bottleneck, heo2021rethinking, graham2021levit, el2021xcit} have shown impressive results in vision tasks. Recent studies have shown that transformer-based architectures renew the state of the art across a wide range of subject areas, including image classification\cite{dosovitskiy2020image, touvron2020training}, object detection and semantic segmentation\cite{carion2020end, liu2021swin, wang2021pyramid, zhang2021multi, zheng2021rethinking}, and generative models\cite{jiang2021transgan, esser2021taming, durall2021combining}.

Despite its great successes, the original self-attention (SA) mechanism has $O(N^2)$ time and memory complexity due to the matrix multiplication of $\sigma (QK^T)\in R^{N\times N}$ and $V$. This is one of the well-known drawbacks of traditional transformers. For vision tasks, $N$ is proportional to the input resolution. This means that SA consumes 16 times the computational resources if the width and height of the input image are doubled.

Here, we propose a new model that implements an alternative novel SA mechanism to avoid this drawback. It is called the \textbf{U}nit \textbf{F}orce \textbf{O}perated \textbf{Vi}sion \textbf{T}ransformer, or UFO-ViT. Our key method replaces softmax nonlinearity with a simple $L_2$-norm. Using the associative law of matrix multiplication, our new SA algorithm requires much less computational resources than the original SA.
\input{table/params_acc_comparison}

Although similar approaches have been proposed by Performer\cite{heo2021rethinking} and Efficient Attention\cite{shen2021efficient}, El et al. have shown that these approaches cause performance degradation for XCiT\cite{el2021xcit}. Unlike the previous results, UFO-ViT achieves higher or competitive results on the benchmark datasets of vision tasks compared with state-of-the-art models. This is detailed in Section \ref{sec:experiments}, including the results on the ImageNet1k\cite{deng2009imagenet} and COCO benchmarks\cite{lin2014microsoft}. These results were obtained without any extra-large datasets, such as ImageNet21k, or distilled knowledge from another model.

The main contributions in this paper are summarized as follows:

1. We propose a novel \textit{constraint} scheme, XNorm, that generates a unit hypersphere to extract relational features. As detailed in Section \ref{sec:methods}, this scheme prevents SA from being dependent on initialization. Furthermore, it eliminates non-linearity from SA by replacing the softmax function. Our module has $\text{O}(N)$ complexity, and it handles high-resolution inputs efficiently.

2. We demonstrate that UFO-ViT can be adopted for general purposes. Our models are tested on both image classification and dense prediction tasks. While our proposed method has linear complexity, the UFO-ViT models outperform most of the state-of-the-art models based on transformers at lower capacity and FLOPS. In particular, our models perform well in lightweight regimes.

3. We empirically show that UFO-ViT models have faster inference speed and require less GPU memory. For various resolutions, the required computational resources were not significantly increased. Also, the weights used in our model are irrelevant to the resolution. This is a useful characteristic for dense prediction tasks such as object detection and semantic segmentation. Most of dense prediction tasks require a higher resolution than does the pre-training stage, that is, MLP-based structures\cite{tolstikhin2021mlp, touvron2021resmlp} require additional post-processing to adjust to various resolutions.

%% file: table/params_acc_comparison.tex
\begin{figure}[hbt!]
\begin{center}
\includegraphics[width=\linewidth]{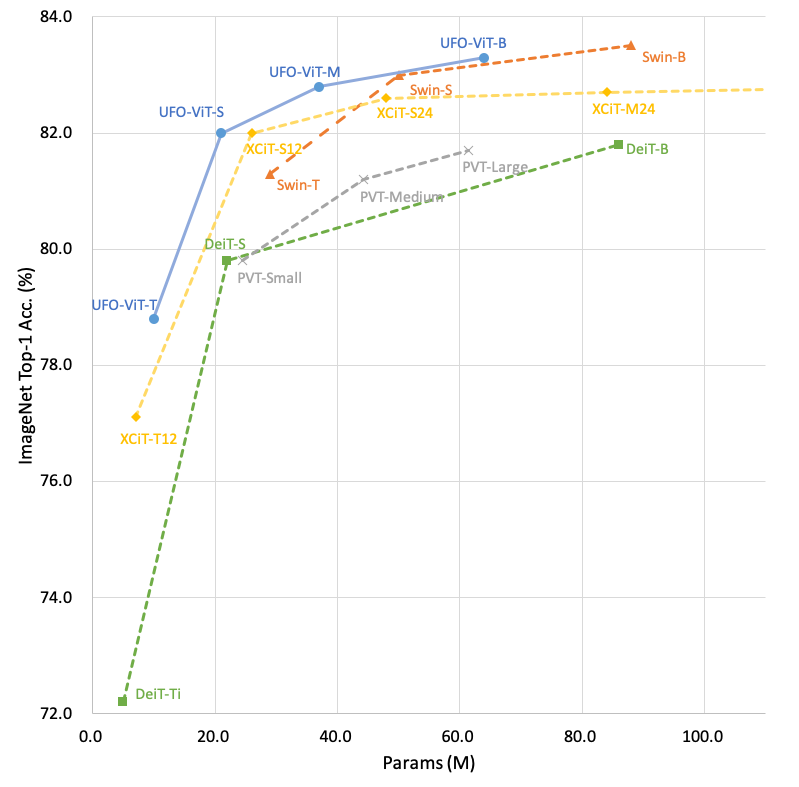}
\end{center}
\caption{\textbf{Top-1 accuracy vs. Model capacity.} Comparison of ImageNet1k top-1 accuracy of various models according to model capacity. Our models show the best results at the same parameter sizes compared to the other models.}
\label{fig:params_acc_comparison}
\end{figure}

%% file: section/related_works.tex
\section{Related Works}
\textbf{Vision transformers.} Dosovitskiy et al.\cite{dosovitskiy2020image} proposed a
vision transformer (ViT), which showed that transformer-based models could be used for vision tasks. After the achievements of ViT, DeiT\cite{touvron2020training} introduced data-efficient training strategies for vision transformers with detailed ablation studies. They solved the ViT data efficiency problem successfully, and most of the current transformer-based models follow their schemes.

In further research, various architectures based on transformer variants have been presented. Touvron et al.\cite{touvron2021going} proposed two simple types of modules. One is the class attention module, which is the additional SA layer used to extract class information. These layers help the model aggregate features from the last outputs. The other is the LayerScale modules. These are learnable parameters for scaling residual connections. This prevents larger models from being overfitted. A simple variant of the LayerScale was presented at ResMLP\cite{touvron2021resmlp}. It is called Affine and LayerScale with a bias term. While MLP-based models are irrelevant to our model, we apply Affine modules to our model as scalers.

\textbf{Spatial features.} As CNN-based studies have shown, human heuristics for extracting spatial structures work well for vision tasks. Several studies have suggested their own methods instead of adopting prior works. Liu et al.\cite{liu2021swin} proposed a shifting window and patch merging. This generates local attention using two types of windows: Normal windows and shifted windows. At the end of each stage, this method merges the patches to preserve large receptive fields without heavy computation. Tokens-to-token ViT, introduced by Yuan et al.\cite{yuan2021tokens}, aims to achieve a similar objective through different approaches. They presented a method of overlapping tokens to locally correlate patches. They did not use additional methods to reduce the computation, except when using small channels. Zhang et al.\cite{zhang2021aggregating} proposed a nested SA structure. They split patches into several blocks and applied local attention merging to the blocks using convolutional pooling layers. Chen et al. used features from small and large patches, implementing their own cross-attention structure\cite{chen2021crossvit}. In a similar vein, an alternative SA method was introduced by Chu et al. \cite{chu2021twins}. Their model used global attention and local attention to handle both types of spatial features.

\textbf{Hybrid architectures.} Various methods for integrating convolutional layers\cite{heo2021rethinking, wang2021pyramid, graham2021levit, el2021xcit, xiao2021early, hassani2021escaping} instead of searching for new spatial structures have been introduced. LeViT, designed by Graham et al.\cite{graham2021levit}, applies multi-stage networks to transformers using SA with convolution and pooling layers. Xiao et al.\cite{xiao2021early} found that replacing linear patch embedding layers with convolutions helps transformers better capture low-level features. This is very similar to the stemming stage of existing CNN networks. El-Nouby et al. introduced local patch interactions in XCiT\cite{el2021xcit}. With two depthwise convolutions\cite{chollet2017xception} added after XCA, XCiT achieved better performance.

Our models are generally inspired by the intrinsic optimization strategies that XCiT\cite{el2021xcit} introduced, while we present our own SA method.

\textbf{Efficient self-attention.} Instead of architectural strategies, many approaches have been proposed to solve the $\text{O}(N^2)$ problem of the SA mechanism. They are summarized in several categories: those that use their own spatial patterns\cite{ho2019axial, child2019generating, sukhbaatar2019adaptive}, those that use various low-rank factorization methods \cite{choromanski2020rethinking, shen2021efficient, wang2020linformer}, those that use linear approximation by sampling important tokens\cite{kitaev2020reformer, xiong2021nystr}, and those that use cross-covariance matrices instead of Gram matrices\cite{el2021xcit}. Although detailed methods are quite different, our UFO scheme is mainly related to low-rank factorization methods.

%% file: section/methods.tex
\section{Methods} \label{sec:methods}
\input{table/ufo_figure}
\input{table/ufo_scheme}
The structure of our model is shown in Figure \ref{fig:ufo-vit}. It is a mixture of
convolutional layers, a UFO module, and a simple feed-forward MLP layer. In this section, we discuss how our proposed method can replace the softmax function and ensure linear complexity.
\input{articles/theory}

\input{articles/XNorm}
\input{articles/ufo-vit}

%% file: table/ufo_figure.tex
\begin{figure}[t] 
\begin{center}
\includegraphics[width=0.7\linewidth]{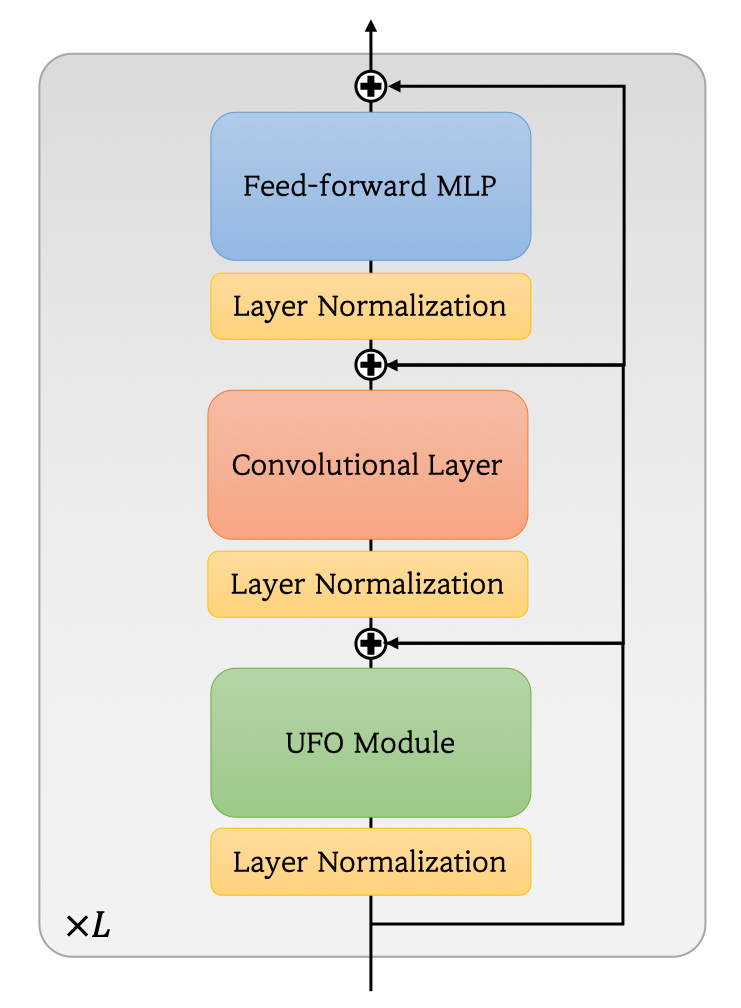}
\end{center}
\caption{\textbf{Overview of UFO-ViT module.} Note that affine layers\cite{touvron2021resmlp} are following each module.} \label{fig:ufo-vit}
\end{figure}

%% file: table/ufo_scheme.tex
\begin{figure}[t] 
\begin{center}
\includegraphics[width=\linewidth]{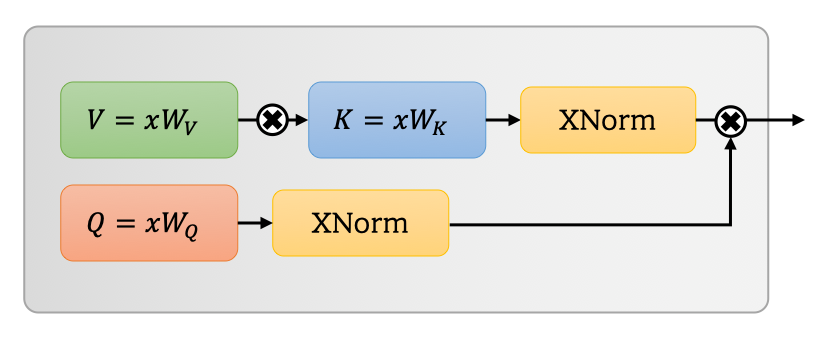}
\end{center}
\caption{\textbf{UFO module.}} \label{fig:ufo_module}
\end{figure}

%% file: articles/theory.tex
\subsection{Basic Structure} \label{sec:theory}
For an input $\textbf{x} \in \mathbb{R}^{N\times C}$, the traditional SA mechanism is formulated as follows:
\begin{align}\label{eq:sa}
&A(\textbf{x})=\sigma(QK^T/\sqrt{d_k})V\\
&Q=\textbf{x}W_Q, K=\textbf{x}W_K, V=\textbf{x}W_V
\end{align}
where $A$ denotes the attention operator.

$\sigma(QK^{T})V$ can be decomposed into $\text{O}(N\times h+h\times N)$ if the nonlinearity of softmax is eliminated. Our approach uses XNorm instead of softmax, which allows SA modules to compute $K^TV$ first. Because applying identity causes degradation, we suggest a simple constraint to prevent it.

Our proposed method, called \textbf{cross-normalization} or \textbf{XNorm}, is defined as follows:
\begin{align}\label{eq:ufo} &A(\textbf{x})=\text{XN}_{\text{dim}=\text{filter}}(Q)(\text{XN}_{\text{dim}=\text{s pace}}(K^{T}V))\\
&\text{XN}(\textbf{a})\coloneqq \cfrac{\gamma\textbf{a}}{\sqrt{\sum_{i=0}^{h} ||\textbf{a}||^2}}
\end{align}
where $\gamma$ is a learnable parameter and $h$ is the number of embedding dimensions. It is a common $L_2$-norm, but it is applied along two dimensions: the spatial dimension of $K^{T}V$ and the channel dimension of $Q$. Thus, it is called \textit{cross}-normalization.

Using the associative law, the key and value are multiplied first, and the query is multiplied afterward. This is depicted in Figure \ref{fig:ufo_module}. Both multiplication operations have a complexity of $\text{O}(hNd)$, so this process is linear to $N$.

%% file: articles/XNorm.tex
\input{table/model_setup}
\subsection{XNorm}
\input{table/centroids_figure}

XNorm is applied to both query and output.
\begin{align}\label{eq:qkv}
&A(\textbf{x})=
\begin{bmatrix}
\hat{q}_{0}\cdot\hat{k}_{0} & \hat{q}_{0}\cdot\hat{k}_{1} & \cdots & \hat{q}_{0}\cdot\hat{k}_{h}\\
\hat{q}_{1}\cdot\hat{k}_{0} & \hat{q}_{1}\cdot\hat{k}_{1} & \cdots & \hat{q}_{1}\cdot\hat{k}_{h}\\
\vdots & \vdots & \ddots & \vdots\\
\hat{q}_{N}\cdot\hat{k}_{0} & \hat{q}_{N}\cdot\hat{k}_{1} & \cdots & \hat{q}_{N}\cdot\hat{k}_{h}
\end{bmatrix}\\
&\hat{q}_i=\text{XN}[(Q_{i0}, Q_{i1}, \cdots, Q_{ih})]\\ &\hat{k}_i=\text{XN}[([K^{T}V]_{0i}, [K^{T}V]_{1i}, \cdots, [K^{T}V]_{hi})] \end{align}
where $\textbf{x}$ denotes the input. Finally, the projection weight scales and aggregates the dot product terms using a weighted sum.
\begin{align}\label{eq:proj}
[W_{\text{proj}}A(\textbf{x})]_{ij}=\sum^h_{m=1} w_{mj}\hat{q}_i\cdot\hat{k}_j \end{align}

In this formulation, relational features are defined by the cosine similarity between patches and clusters. XNorm restricts the features of each pixel in the query and clusters to unit vectors. This prevents their values from suppressing relational properties by regularizing them to a limited length. If they have arbitrary values, the region of attention is dependent on the initialization.

However, this interpretation is not sufficient to explain \textit{why} XNorm has to be the $L_2$-normalization form only. Here, we introduce another theoretical view that considers a simple physical simulation.

\textbf{Details of XNorm.} Considering the residual connections, the output of an arbitrary module is formulated as follows:
\begin{align}\label{eq:res}
\textbf{x}_{n+1}=\textbf{x}_{n}+f(\textbf{x}_n)
\end{align}
where $n$ and $x$ denote the index of the current layer and the input image, respectively. If we assume that \textbf{x} is the displacement of a certain object and $n$ is time, then the above equation can be re-defined as:
\begin{align}\label{eq:res_t}
&\textbf{x}_{t+1}=\textbf{x}_{t}+f(\textbf{x}_t)\\
&f(\textbf{x})=\cfrac{\Delta \textbf{x}}{\Delta t}
\end{align}

Most neural networks are discrete, so $\Delta t$ is constant. (Let $\Delta t=1$ for simplicity.) The residual term expresses velocity, so this term represents the \textit{force} term if the particle has a unit mass and $\Delta t=1$.

In physics, Hooke's law is defined as the dot product of the elasticity vector $k$ and the displacement vector $x$. The elastic force generates harmonic potential $U$, a function of $x^2$. Physically, the potential energy interferes with the path of the particle moving through it. (Imagine a ball moving around a track with a parabolic shape.)
\begin{align} \textbf{F}=-\textbf{k}\cdot\textbf{x}\label{eq:hooke}\\ U=\cfrac{1}{2}\:kx^2\label{eq:potential}
\end{align}

The above formulation is generally used to approximate the potential energy of a molecule at $x\approx0$. For multiple molecules, the linearity of elasticity can be used: \begin{align}\label{eq:multi_hooke} \textbf{F}=-\sum^n_{i=1}\textbf{k}_i\cdot\textbf{x}
\end{align}

This equation is similar to Eq.\ref{eq:proj}, except that the above formula is not normalized. (Note that the $w_{mj}$ terms are \textit{static}. They are irrelevant to each batch, so they do not perform significant roles in this simulation.) Assume that a small number of $\textbf{k}$ values are too large. The shape of their potentials is wide and deep. (See Eq.\ref{eq:potential}.) If a certain particle moves around them, it cannot easily escape. In this case, only two outcomes can occur: 1. Collapsing case. The particles are aligned in the direction of the largest $\textbf{k}$. 2. The relational features between particles are neglected. A few particles with large $\textbf{x}$ survive collapsing.

This case is exactly the same as in the previous section. XNorm forces all vectors to be unit vectors to prevent this situation. In other words, XNorm is not a \textit{normalization}, but a \textit{constraint}. This is why our method is called a unit force operation, or simply UFO.

To show this empirically, we demonstrate that other normalization methods cannot work well. For detailed results, refer to our ablation study. (See Table \ref{tb:imagenet_ab}.)

\textbf{Feed-forward Networks (FFN).} In the attention module, FFN cannot be ignored. However, this is a simple interpretation. The FFN is static and is not dependent on the spatial dimension. Physically, this type of function represents the \textit{driven force} of the harmonic oscillator equation.
\begin{align}\label{eq:driven_force}
\textbf{F}=-\textbf{k}\cdot\textbf{x}+g(t) \end{align}

This increases or reduces the amplitude physically. In other words, it boosts or kills features that are irrelevant to spatial relations.

%% file: table/model_setup.tex
\begin{table*}[hbt!]
\centering
\begin{tabular}{c|c|c|c|c|c|c|c|c}
\hline
\rowcolor[HTML]{EFEFEF} 
Model & Depth & \#Dim & \#Embed & \#Head & GFLOPS & Params (M) & Res. & Patch Size \\ \hline
UFO-ViT-T & 24 & 192 & 96 & 4 & 1.9  & 10.0 & 224 & 16 \\ \hline
UFO-ViT-S & 12 & 384 & 128 & 8 & 3.7  & 20.7 & 224 & 16 \\ \hline
UFO-ViT-M & 24 & 384 & 128 & 8 & 7.0  & 37.3 & 224 & 16 \\ \hline
UFO-ViT-B & 24 & 512 & 128 & 8 & 11.9 & 63.7 & 224 & 16 \\ \hline
\end{tabular}
\caption{\textbf{UFO-ViT models.}} \label{tb:model_setup}
\end{table*}

%% file: table/centroids_figure.tex
\begin{figure}[t] 
\begin{center}
\includegraphics[width=0.7\linewidth]{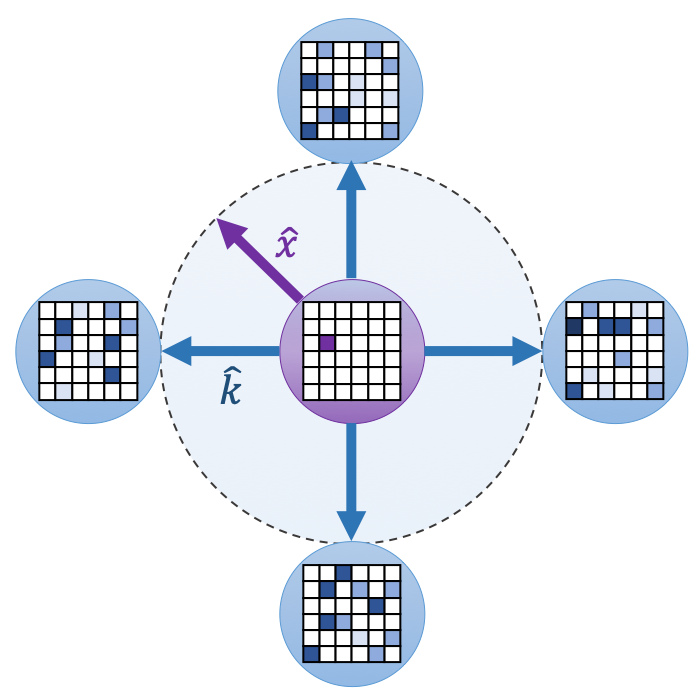}
\end{center}
\caption{\textbf{Centroids generating harmonic potentials.} Each patch, represented as a particle, is interfered by parabolic potentials generated from $h$ clusters.}
\label{fig:centroids_figure}
\end{figure}
\textbf{Replace softmax to XNorm.} In XNorm, key and value of self-attention multiplied directly. It generates $h$ clusters through linear kernel method.
\begin{align}\label{eq:kv}
[K^{T}V]_{ij}=\sum^n_{k=1}K^T_{ik}V_{kj}
\end{align}

%% file: articles/ufo-vit.tex
\subsection{UFO-ViT}\label{sec:ufo-vit}
To build our UFO-ViT model, we adopted architectural strategies from earlier vision transformer models\cite{graham2021levit, xiao2021early, el2021xcit, touvron2021going}. In this section, we introduce several intrinsic structures that improve performance. The overall structure is illustrated in Figure \ref{fig:ufo-vit}.

\textbf{Replace linear patch embedding with convolutions.} Several recent studies \cite{graham2021levit, xiao2021early} claimed that early convolutional layers help vision transformers to be well-trained. To adopt their strategy, we used convolutional layers instead of linear patch-embedding layers.

\textbf{Positional encoding.} We used positional encoding as a learnable parameter. This was proposed in the original vision transformer\cite{dosovitskiy2020image}. For smaller or larger resolutions, we resized the positional encoding with bicubic interpolation.

\textbf{Multi-headed attention.} Following the original transformer\cite{vaswani2017attention}, our modules are multi-headed for better regularization. The $\gamma$ parameter in Eq.\ref{eq:ufo} is applied to all heads to scale the importance of each head.

\textbf{Convolutional layers.} Designing an extra module to extract local features is not a new idea. We chose the most simplistic method by adding various types of convolutional layers. We experimented with both the simple depthwise convolutions and the local patch interaction (LPI) layers proposed in XCiT\cite{el2021xcit}. We found that the former showed better performance on the regimes overall.

\textbf{Class attention.} In the ImageNet1k experiments, we used the class attention layers presented in CaiT\cite{touvron2021going}. This helps the class token gather spatial information. Class attention is computed on class token only to reduce computation, as in the original paper. We implemented the class attention layers using UFO modules, whereas CaiT used the SA module for class attention.

%% file: section/experiments.tex
\section{Experiments} \label{sec:experiments}
\input{table/hyperparams}
\input{articles/imagenet1k}
\input{articles/coco}
\input{articles/computational_efficiency}

%% file: table/hyperparams.tex
\begin{table*}[hbt!]
\centering
\begin{tabular}{|l|c|c|c|c|}
\hline
\rowcolor[HTML]{EFEFEF} 
\multicolumn{1}{|c|}{\cellcolor[HTML]{EFEFEF}}                                 & \multicolumn{4}{c|}{\cellcolor[HTML]{EFEFEF}Models} \\ \cline{2-5} 
\rowcolor[HTML]{EFEFEF} 
\multicolumn{1}{|c|}{\multirow{-2}{*}{\cellcolor[HTML]{EFEFEF}Hyperparameter}} & UFO-ViT-T & UFO-ViT-S       & UFO-ViT-M       & UFO-ViT-B       \\ \hline
learning rate     & 5e-4 & 5e-4  & \multicolumn{2}{c|}{4e-4} \\ \hline
DropPath\cite{huang2016deep}         & 0.05 & 0.1   & 0.15        & 0.25        \\ \hline
RandAugment\cite{cubuk2020randaugment}      & 2, 7 & 2, 7 & 2, 9       & 3, 12       \\ \hline
\end{tabular}
\caption{\textbf{Hyperparameters for image classification.} All the other hyperparameters are same as DeiT\cite{touvron2020training}.} \label{tb:hyperparams}
\end{table*}

%% file: articles/imagenet1k.tex
\subsection{Image Classification}
\textbf{Dataset.} For the image classification task, we evaluated our models using the ImageNet1k\cite{deng2009imagenet} dataset. No extra dataset or distilled knowledge from a convolutional teacher were used.

\textbf{Implementation details.} Our setup was almost the same as that of DeiT\cite{touvron2020training}. However, we optimized some hyperparameters according to the model size. For details, see Table \ref{tb:hyperparams}. The learning rate followed the linear scaling rule\cite{you2017large} and was scaled per the 512 batch size. We trained our model for 400 epochs using the AdamW optimizer\cite{loshchilov2017decoupled}.

The learning rate was linearly warmed up during the first five epochs and decayed with a cosine schedule thereafter. RandAugment\cite{cubuk2020randaugment} was used for data augmentation. As shown in Table \ref{tb:hyperparams}, we applied a stronger augmentation in larger models.
\input{table/flops_acc_comparison}

\textbf{Ablation study.} Our ablation study mainly focused on the importance of XNorm and architectural optimizations, as explained in Section \ref{sec:ufo-vit}. All experiments were performed using the UFO-ViT-M model. We show that most of the other normalization methods failed to train. In other words, they did not properly decrease the loss. This can serve as implicit evidence to prove that our theoretical interpretation is reasonable. Interestingly, the application of a single $L_2$-norm also shows poor performance. All the results are presented in Table \ref{tb:imagenet_ab}.
\input{table/imagenet_ablation}

\textbf{Fine-tune at higher resolution.} Instead of training from scratch again, we fine-tuned UFO-ViT-M and UFO-ViT-B at the higher resolution. During fine-tuning stage, the heads of models are updated only. This method prevented the models from over-fitting and made the training faster. Our models could achieve the higher performance in 0.1$\times$ training time compared to learning from scratch.

\textbf{Comparison with state-of-the-art models.} We experimented with four models that used the same architectural design schemes as DeiT\cite{touvron2020training}. (See Table \ref{tb:imagenet_comparison}.) As summarized in Figure \ref{fig:params_acc_comparison} and \ref{fig:flops_acc_comparison}, all our models showed better performance and computational efficiency than most of the concurrent transformer-based models.
\input{table/visualized_attention}

\textbf{Visualization of attention maps.} We visualized the attention maps on the class token of the UFO-ViT-S model. Because they could not be computed directly, a pseudo-inverse algorithm was used to approximate them. All samples were randomly sampled in the validation set.

Figure \ref{fig:vis_attn} qualitatively shows that our model successfully captures perception information using SA. Each attention map attends well to the main objects, even if the sample has a complicated shape. Those results also support our theoretical assumptions in section \ref{sec:theory}.
\input{table/comparison_imagenet1k}

%% file: table/flops_acc_comparison.tex
\begin{figure}[hbt!]
\begin{center}
\includegraphics[width=\linewidth]{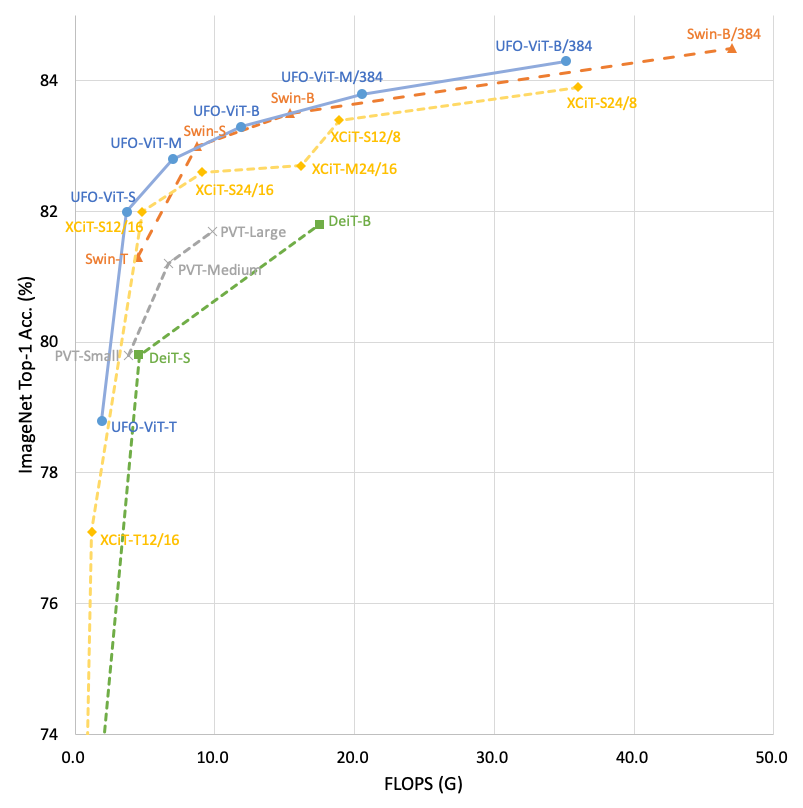}
\end{center}
\caption{\textbf{Top-1 accuracy vs. FLOPS.} Comparison of ImageNet1k top-1 accuracy of various models according to FLOPS. The results of models fine-tuned at higher resolutions are added. Our models show the best results at the same FLOPS compared to the other models.}
\label{fig:flops_acc_comparison}
\end{figure}

%% file: table/imagenet_ablation.tex
\begin{table}[hbt!]
\centering
\begin{tabular}{l|c}
\hline
\rowcolor[HTML]{EFEFEF} 
Method              & Top-1 Acc. (\%) \\ \hline
Baseline(Linear Embed+XNorm)            & 81.8                       \\ \hline
XNorm $\rightarrow$ LN\cite{ba2016layer}, GN\cite{wu2018group}, IN\cite{ulyanov2016instance}          & Failed                     \\ \hline
XNorm $\rightarrow$ Learnable $p$-Norm & 81.8                       \\ \hline
XNorm $\rightarrow$ Single $L_2$-norm & Failed                       \\ \hline
Linear Embed\cite{dosovitskiy2020image} $\rightarrow$ Conv Embed         & 82.0                       \\ \hline
Hyperparameter Optimization           & 82.8                       \\ \hline
\end{tabular}
\caption{\textbf{Ablation study.} The results of ablation study on ImageNet1k classification for UFO-ViT-M model. Note that single $L_2$-Norm means applying $L_2$-Norm to only one of query and key-value interaction.} \label{tb:imagenet_ab}
\end{table}

%% file: table/visualized_attention.tex
\begin{figure}[hbt!]
\begin{center}
\includegraphics[width=\linewidth]{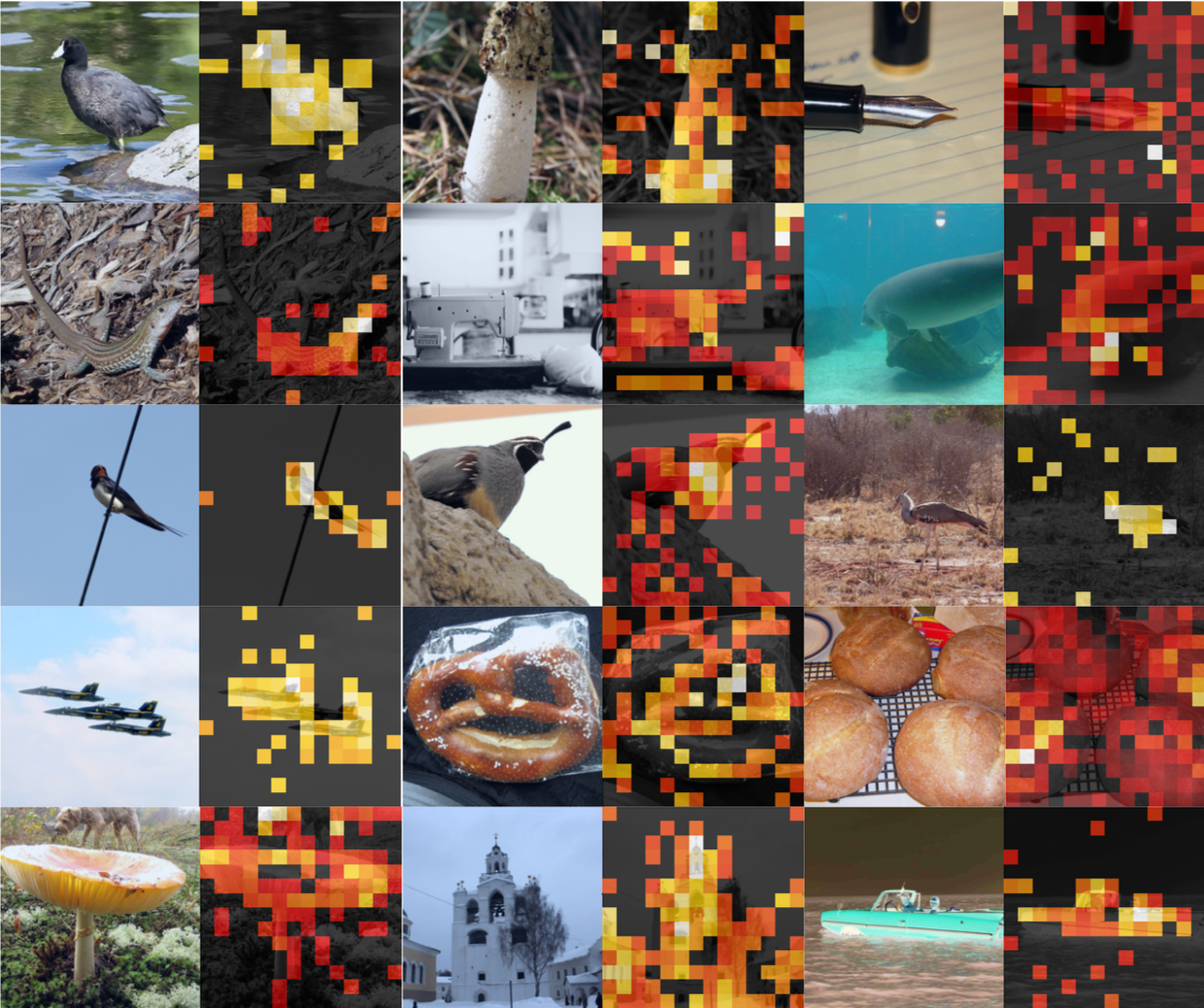}
\end{center}
\caption{\textbf{Visualized attentions.} Visualization of attention matrices using pseudo-inverse scheme. These matrices are extracted from class attention module of pretrained UFO-ViT-S. The brighter colors represent the values closer to 1. The thresholds are adjusted manually.}
\label{fig:vis_attn}
\end{figure}

%% file: table/comparison_imagenet1k.tex
\begin{table}[hbt!]
\centering
\begin{tabular}{l|c|c|c}
\hline
\multirow{2}{*}{Model} & Top-1 & Params & FLOPs \\ 
 & Acc. & (M) & (G) \\ \hline
RegNetY-1.6G\cite{radosavovic2020designing} & 78.0 & 11 & 1.6 \\
DeiT-Ti\cite{touvron2020training} & 72.2 & 5 & 1.3 \\
XCiT-T12/16\cite{el2021xcit} & 77.1 & 26 & 1.2 \\
\rowcolor[HTML]{EFEFEF}
UFO-ViT-T & 78.8 & 10 & 1.9 \\ \hline

ResNet-50\cite{he2016deep} & 75.3 & 26 & 3.8 \\
RegNetY-4G\cite{radosavovic2020designing} & 80.0 & 21 & 4.0 \\
DeiT-S\cite{touvron2020training} & 79.8 & 22 & 4.6 \\
Swin-T\cite{liu2021swin} & 81.3 & 29 & 4.5 \\
XCiT-S12/16\cite{el2021xcit} & 82.0 & 26 & 4.8 \\
\rowcolor[HTML]{EFEFEF}
UFO-ViT-S & 82.0 & 21 & 3.7 \\ \hline

ResNet-101\cite{he2016deep} & 75.3 & 47 & 7.6 \\
RegNetY-8G\cite{radosavovic2020designing} & 81.7 & 39 & 8.0 \\
Swin-S\cite{liu2021swin} & 83.0 & 50 & 8.7 \\
XCiT-S24/16\cite{el2021xcit} & 82.6  & 48 & 9.1 \\
\rowcolor[HTML]{EFEFEF}
UFO-ViT-M & 82.8 & 37 & 7.0 \\ \hline

RegNetY-16G\cite{radosavovic2020designing} & 82.9 & 84 & 16.0 \\
DeiT-B\cite{touvron2020training} & 81.8 & 86 & 17.5 \\
Swin-B\cite{liu2021swin} & 83.5 & 88 & 15.4 \\
XCiT-M24\cite{el2021xcit} & 82.9 & 84 & 16.2 \\
\rowcolor[HTML]{EFEFEF}
UFO-ViT-B & 83.3 & 64 & 11.9 \\ \hline

EfficientNet-B7\cite{tan2019efficientnet} & 84.3 & 66 & 37.0 \\
XCiT-S24/8\cite{el2021xcit} & 83.9 & 48 & 36.0 \\
Swin-B/384\cite{liu2021swin} & 84.5 & 48 & 47.0 \\
\rowcolor[HTML]{EFEFEF}
UFO-ViT-M/384 & 83.8 & 37 & 20.5 \\
\rowcolor[HTML]{EFEFEF}
UFO-ViT-B/384 & 84.3 & 64 & 35.1 \\ \hline
\end{tabular}
\caption{\textbf{Comparison with the state of the art models.} The image classification results, model capacity, and FLOPS of various models on ImageNet1k dataset.} \label{tb:imagenet_comparison}
\end{table}

%% file: articles/coco.tex
\subsection{Object Detection with Mask R-CNN}
\textbf{Implementation details.} Our models were evaluated on the COCO benchmark dataset\cite{lin2014microsoft} for the object detection task. We used UFO-ViT as the backbone and mask R-CNN\cite{he2017mask} as the detector. This implementation was based on the MMDetection library.\cite{chen2019mmdetection} The scheduling and data augmentation follow the setup of DETR\cite{carion2020end}. We used 16 NVIDIA A100 GPUs for training over 36 epochs with two batch sizes per GPU using the AdamW optimizer. Unlike for image classification, we used the same hyperparameters for all models. All experiments were performed on a 3x schedule. The input resolution was fixed at $800\times1333$ for all the experiments.

\textbf{Evaluation on COCO dataset.} We compared CNNs\cite{he2016deep, xie2017aggregated} and transformer-based vision models on object detection and instance segmentation tasks. To make the comparison fair, the experimental environment was the same for all the results. All models were pre-trained on the ImageNet1k dataset.

According to Table \ref{tb:coco}, our models significantly outperform the CNN-based models. In addition, they achieve higher or more competitive results than do state-of-the-art vision transformers.
\input{table/coco_mAP_table}

The modules in XCiT\cite{el2021xcit} models have a structure similar to that of the UFO modules except for the SA scheme. Hence, they show a mAP curve similar to that of UFO-ViT according to the model capacity. They perform slightly better than our model in a similar design space. It is possible that the XCiT models have a larger embedding space $d$.

Swin transformer\cite{liu2021swin} models showed better results in the overall regime. We infer that this is because their architectural strategy is better optimized for dense prediction tasks. Notably, UFO-ViT-B performs slightly worse than UFO-ViT-M on the bounding box detection task, but slightly better in detecting smaller bounding boxes and overall instance segmentation scores.

%% file: table/coco_mAP_table.tex
\begin{table*}[hbt!]
\centering
\begin{tabular}{c|c|c|c|c|c|c|c}
\hline
\rowcolor[HTML]{EFEFEF} 
Backbone &
  Params (M) &
  $\text{AP}^b$ &
  $\text{AP}^b_{50}$ &
  $\text{AP}^b_{75}$ &
  $\text{AP}^m$ &
  $\text{AP}^m_{50}$ &
  $\text{AP}^m_{75}$ \\ \hline
ResNet50\cite{he2016deep}   & 44.2  & 41.0 & 61.7 & 44.9 & 37.1 & 58.4 & 40.1 \\
PVT-Small\cite{wang2021pyramid}  & 44.1  & 43.0 & 65.3 & 46.9 & 39.9 & 62.5 & 42.8 \\
Swin-T\cite{liu2021swin}     & 47.8  & 46.0 & 68.1 & 50.3 & 41.6 & 65.1 & 44.9 \\
XCiT-S12/16\cite{el2021xcit}   & 44.3  & 45.3 & 67.0 & 49.5 & 40.8 & 64.0 & 43.8 \\
\rowcolor[HTML]{EFEFEF} 
UFO-ViT-S  & 39.7  & 44.6 & 66.7 & 48.7 & 40.4 & 63.6 & 42.9 \\ \hline
ResNet101\cite{he2016deep}  & 63.2  & 42.8 & 63.2 & 47.1 & 39.2 & 60.1 & 41.3 \\
PVT-Medium\cite{wang2021pyramid} & 63.9  & 44.2 & 66.0 & 48.2 & 40.5 & 63.1 & 43.5 \\
Swin-S\cite{liu2021swin}     & 69.0  & 48.5 & 70.2 & 53.5 & 43.3 & 67.3 & 46.6 \\
XCiT-S24/16\cite{el2021xcit}   & 65.8  & 46.5 & 68.0 & 50.9 & 41.8 & 65.2 & 45.0 \\
\rowcolor[HTML]{EFEFEF} 
UFO-ViT-M  & 56.4  & 46.0 & 68.2 & 50.0 & 41.0 & 64.6 & 43.7 \\ \hline
ResNeXt101-64\cite{xie2017aggregated}  & 101.9  & 44.4 & 64.9 & 48.8 & 39.7 & 61.9 & 42.6 \\
PVT-Large\cite{wang2021pyramid}  & 81.0  & 44.5 & 66.0 & 48.3 & 40.7 & 63.4 & 43.7 \\
XCiT-M24/16\cite{el2021xcit}   & 101.1 & 46.7 & 68.2 & 51.1 & 42.0 & 65.6 & 44.9 \\
\rowcolor[HTML]{EFEFEF} 
UFO-ViT-B  & 82.4  &    45.8  & 67.4     & 50.1     & 41.2     & 64.5 & 44.1    \\ \hline
\end{tabular}
\caption{\textbf{Object detection performance on the COCO val2017.}}\label{tb:coco}
\end{table*}

%% file: articles/computational_efficiency.tex
\input{table/available_batch_size}
\input{table/memory_consumption}
\input{table/resolution_throughput_comparison}
\subsection{Measuring Computational Efficiency}
We measured the various computational resources required for the inference. All measurements were performed on a single V100 GPU with 16 GB of VRAM.

\textbf{Memory efficiency.} As shown in Figure \ref{fig:memory_consumption}, we determined that our models consumed much less memory for larger resolutions compared to DeiT models\cite{touvron2020training} and Swin Transformers\cite{liu2021swin}. Our model can process up to a 4$\times$ batch size compared with other models showing similar performance. (See Figure \ref{fig:available_batch_size}).

\textbf{GPU throughput.} Our model is faster than other models showing similar performance. (See Figure \ref{fig:resolution_throuput_comparison}.)
In addition, the GPU throughput of our model is significantly lower than that of DeiT\cite{touvron2020training} for higher input resolution.

%% file: table/available_batch_size.tex
\begin{figure}[hbt!]
\begin{center}
\includegraphics[width=\linewidth]{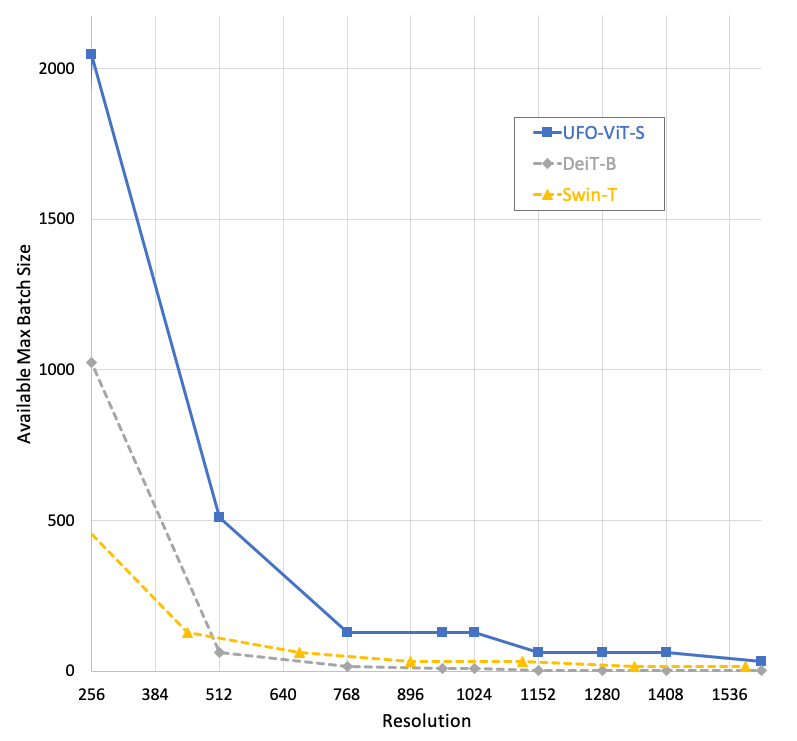}
\end{center}
\caption{\textbf{Maximum batch size on single GPU.} Comparison of maximum available batch size on single GPU. We scaled batch size by 2 until OOM causes.}
\label{fig:available_batch_size}
\end{figure}

%% file: table/memory_consumption.tex
\begin{figure}[hbt!]
\begin{center}
\includegraphics[width=\linewidth]{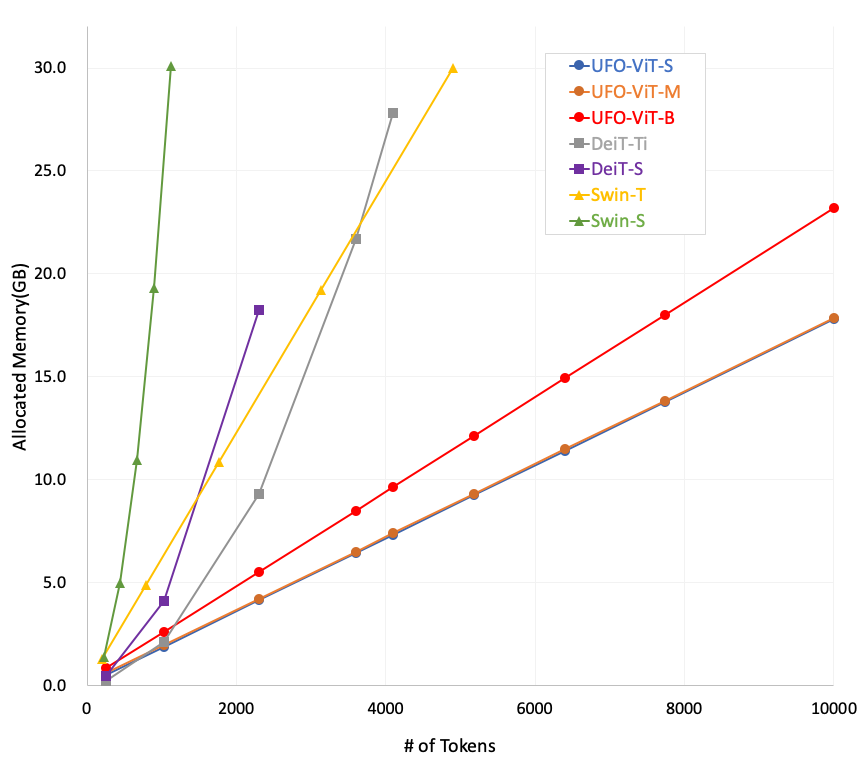}
\end{center}
\caption{\textbf{Allocated memory vs. \# of tokens.} To check the linearity of our models empirically, we measured the maximum value of allocated GPU memory on different resolutions. For a batch size of 64, the memory consumption of our models showed linearity with the number of tokens. Moreover, our models required significantly less memory than the other models.}
\label{fig:memory_consumption}
\end{figure}

%% file: table/resolution_throughput_comparison.tex
\begin{figure}[hbt!]
\begin{center}
\includegraphics[width=\linewidth]{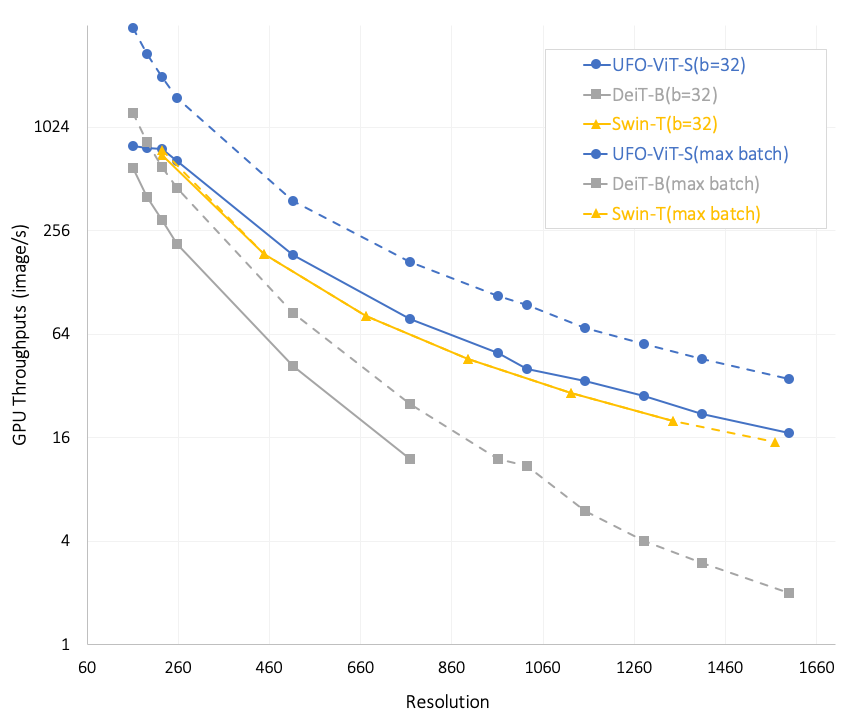}
\end{center}
\caption{\textbf{GPU throughput according to the input resolution.} Note that the scale of throughput axis is $\log _4$ scale. 'max batch' means throughput measured on maximum available batch size.}
\label{fig:resolution_throuput_comparison}
\end{figure}

%% file: section/conclusion.tex
\section{Conclusion}
In this paper, we proposed a simple method that ensures linear complexity for SA without
loss of performance. By replacing the softmax function, we removed the quadratic operation using the associative law of matrix multiplication. This type of factorization has typically caused performance degradation in earlier studies. The UFO-ViT models outperformed most of the existing state-of-the-art transformer-based and CNN-based models for image classification. We have shown that our models can also be deployed well for general purposes. Our UFO-ViT models show performance on dense prediction tasks that is competitive with or better than earlier models. With more optimized structures for dense prediction, we expect our models to become more efficient and to perform better.